\documentclass[conference]{IEEEtran}
\IEEEoverridecommandlockouts
\usepackage{graphicx}
\usepackage{tabularx}
\usepackage{amsmath}
\begin{document}

\title{SpikeRL: A Scalable and Energy-efficient Framework for Deep Spiking Reinforcement Learning$^*$
\vspace*{-1ex}
\thanks{$^*$This work was supported by the U.S. Department of Energy,
Office of Science, ASCR under Award Number DE-SC0021419. \\ % FAIR SBI
This material is based upon work supported by the Assistant Secretary of
Defense for Research and Engineering under Air Force Contract No.\
FA8702-15-D-0001. Any opinions, findings, conclusions or recommendations
expressed in this material are those of the author(s) and do not
necessarily reflect the views of the Assistant Secretary of Defense for
Research and Engineering. \\
}
}
% $^1$ EECS, UTK, USA
% \texttt{ttahmid@vols.utk.edu} \\
% $^2$ MIT Lincoln Lab, MA, USA

% \author{\IEEEauthorblockN{Tokey Tahmid$^1$}
% \and
% \IEEEauthorblockN{Mark Gates$^1$}
% \and
% \IEEEauthorblockN{Piotr Luszczek$^{1,2}$}
% \and
% \IEEEauthorblockN{Catherine Schuman$^1$}
% \iffalse
% \IEEEauthorblockA{\textit{Electrical Engineering and Computer Science} \\
% \textit{University of Tennessee, Knoxville}\\
% Knoxville, Tennessee, USA \\
% ttahmid@vols.utk.edu}
% \IEEEauthorblockA{\textit{Electrical Engineering and Computer Science} \\
% \textit{University of Tennessee, Knoxville}\\
% Knoxville, Tennessee, USA \\
% mgates3@icl.utk.edu}
% \IEEEauthorblockA{\textit{Electrical Engineering and Computer Science} \\
% \textit{University of Tennessee, Knoxville}\\
% Knoxville, Tennessee, USA \\
% luszczek@icl.utk.edu}
% \IEEEauthorblockA{\textit{Electrical Engineering and Computer Science} \\
% \textit{University of Tennessee, Knoxville}\\
% Knoxville, Tennessee, USA \\
% cschuman@utk.edu}
% \fi
% }

\author{\IEEEauthorblockN{Tokey Tahmid}
\IEEEauthorblockA{\textit{Department of EECS} \\
\textit{University of Tennessee}\\
Knoxville, TN, USA\\
ttahmid@vols.utk.edu}
\and
\IEEEauthorblockN{Mark Gates}
\IEEEauthorblockA{\textit{Innovative Computing Laboratory} \\
\textit{University of Tennessee}\\
Knoxville, TN, USA\\
mgates3@icl.utk.edu}
\and
\IEEEauthorblockN{Piotr Luszczek}
\IEEEauthorblockA{\textit{MIT Lincoln Lab, MA, USA} \\
\textit{University of Tennessee}\\
Knoxville, TN, USA\\
luszczek@icl.utk.edu}
\and
\IEEEauthorblockN{Catherine D. Schuman}
\IEEEauthorblockA{\textit{Department of EECS} \\
\textit{University of Tennessee}\\
Knoxville, TN, USA\\
cschuman@utk.edu}
}

\maketitle

\begin{abstract}
In this era of AI revolution, massive investments in large-scale data-driven AI systems demand high-performance computing, consuming tremendous energy and resources. This trend raises new challenges in optimizing sustainability without sacrificing scalability or performance. Among the energy-efficient alternatives of the traditional Von Neumann architecture, neuromorphic computing and its Spiking Neural Networks (SNNs) are a promising choice due to their inherent energy efficiency. However, in some real-world application scenarios such as complex continuous control tasks, SNNs often lack the performance optimizations that traditional artificial neural networks have. Researchers have addressed this by combining SNNs with Deep Reinforcement Learning (DeepRL), yet scalability remains unexplored. In this paper, we extend our previous work on SpikeRL, which is a scalable and energy efficient framework for DeepRL-based SNNs for continuous control. In our initial implementation of SpikeRL framework, we depended on the population encoding from the Population-coded Spiking Actor Network (PopSAN) method for our SNN model and implemented distributed training with Message Passing Interface (MPI) through mpi4py. Also, further optimizing our model training by using mixed-precision for parameter updates. In our new SpikeRL framework, we have implemented our own DeepRL-SNN component with population encoding, and distributed training with PyTorch Distributed package with NCCL backend while still optimizing with mixed precision training. Our new SpikeRL implementation is 4.26X faster and 2.25X more energy efficient than state-of-the-art DeepRL-SNN methods. Our proposed SpikeRL framework demonstrates a truly scalable and sustainable solution for complex continuous control tasks in real-world applications.
\end{abstract}


\section{Introduction}

The rapid advancements in Artificial Intelligence (AI) bring about
significant energy costs that pose sustainability challenges. For
instance, OpenAI's GPT-3, a language model with 175 billion parameters, was estimated to consume approximately 1287 MWh for its training and deployment phase~\cite{b27, b34}.
This staggering amount of energy underscores the growing concern as AI systems become more complex and widespread. Future projections are even more alarming, with estimates suggesting that the energy consumption of Nvidia’s new AI servers could surpass the annual energy usage of countries like Argentina and Sweden~\cite{b27, b34}. These figures highlight an urgent need for innovations in energy-efficient computing to mitigate the environmental impact of such technologies while continuing to harness their capabilities.

Spiking Neural Networks (SNNs) are non-von Neumann architecture that is inspired by the human brain  and mimics the neural structure and functioning of the human brain. SNNs offer a compelling energy efficient
alternative to traditional high energy consuming architectures. These networks utilize a non-Von Neumann architecture that inherently supports energy efficiency and parallelism, making them ideal for developing energy efficient AI applications~\cite{b28}. Despite their promise, SNNs have historically struggled to perform at levels comparable to conventional Artificial Neural Networks (ANNs) in real world applications such as complex continuous control tasks, primarily due to their lack of advanced optimization techniques~\cite{b29}.

Recognizing the untapped potential of SNNs, recent research has focused on their integration with Deep Reinforcement Learning (DeepRL) algorithms. This strategic amalgamation aims to harness the biological fidelity and energy efficiency of SNNs alongside the powerful optimization capabilities of DeepRL. Such integration enhances the performance of SNNs in complex control scenarios, where adaptability and precision are paramount.

Building on these concepts, this paper introduces a novel framework named SpikeRL. This framework is specifically designed to leverage the synergies between SNNs and DeepRL, focusing on scalability and efficiency in both training and deployment. It incorporates a DeepRL based SNN with population encoding~\cite{b30} and utilizes the Message Passing Interface (MPI)~\cite{b31} for distributed training across various computational environments. Additionally, it adopts mixed-precision training techniques~\cite{b20} to further optimize computational resource usage.

\section{Related Work}

Neuromorphic computers are non-von Neumann systems composed of artificial neurons and synapses, where the architecture and operational principles are inspired by biological neural networks. Unlike traditional von Neumann computers, which separate processing units (CPUs) from memory units, neuromorphic systems integrate computation and memory. In these systems, neurons process information and synapses store and transmit data, blurring the line between processing and memory elements. The computation in neuromorphic systems is defined by the structure of the neural network and the parameters of these neurons and synapses, rather than explicit instruction sets. Additionally, while conventional computers represent information in binary form, neuromorphic computers use spikes (discrete events) over time to encode information~\cite{b32}.

SNNs offer great potential in the field of robotics~\cite{b1}
including pattern generation, motor control, and navigation. For pattern
generation, Both Cuevas-Arteaga et al.~\cite{b2} and Strohmer et
al.~\cite{b3} employed SNN-based Central Pattern Generators (sCPG) on
the SpiNNaker platform~\cite{b4} to command hexapod robots, for which
they reported achieving the tasks of walking, trotting, or running.
Donati et al.~\cite{b5} and Angelidis et
al.~\cite{b6} demonstrated the performance of SNNs in bio-mimetic
robots, including a lamprey-like model, thus showing the ability to
dynamically control the robots' movement. In terms of motor control,
Dupeyroux et al.~\cite{b7} and Stagsted et al.~\cite{b8} demonstrated
SNN controllers on the Loihi neuromorphic chip for drones. For
navigation, the Spiking RatSLAM project~\cite{b9} and the development of
Gridbot~\cite{b10} used SNNs for spatial awareness and environment
mapping. Bing et al.~\cite{b11} used SNNs for robotic navigation in both
real-world and simulated environments. Lastly, the SDDPG
model~\cite{b12} combined a spiking actor network with a deep critic
network for energy-efficient and map-less navigation.

Building on top of the research of SNNs from robotics, Tang et al.~\cite{b13} introduced a Reinforcement learning integration with SNNs. Their work highlighted the limitations of SNNs in terms of representing information and how this affects learning algorithms. Their proposed population-coded spiking actor network
(PopSAN) incorporated population coding with SNNs to improve their
limitation to represent data for continuous control.
The authors trained and deployed their model on Intel's Loihi
hardware~\cite{b18}, demonstrating a significant reduction in energy
consumption without compromising performance. We extend this work
towards a scalable and efficient infrastructure for deepRL-based SNNs by incorporating MPI and mixed precision.

\emph{Mixed precision training}, particularly the work of Micikevicius et
al.~\cite{b20}, highlights the use of reduced precision arithmetic to
create a balance between computational efficiency and model accuracy.
Using a half-precision format (FP16) instead of single-precision (FP32),
mixed-precision training can significantly lower the memory bandwidth
requirements and speed up computations on Tensor Core units. However,
the narrower dynamic range of FP16 presents potential risks to model
accuracy. The authors counter this by maintaining a master copy of
weights in FP32 and employing loss-scaling to mitigate the
\emph{vanishing gradient problem}, and they perform the FP16 arithmetic
with accumulation in FP32: a mixed-precision approach that maintains the
accuracy of the model~\cite{b35}. We utilize this mixed precision technique in our SpikeRL framework to gain speedup and efficiency without losing performance.

\iffalse
With the knowledge of mixed precision training, distributed training,
and combined architecture of SNN and DeepRL, we move forward to develop our
scalable and efficient SpikeRL infrastructure.
\fi

\section{Spiking Reinforcement Learning (SpikeRL)}

\begin{figure}[tbh]
\centering
% \vspace*{-3ex}
\includegraphics[width=\linewidth]{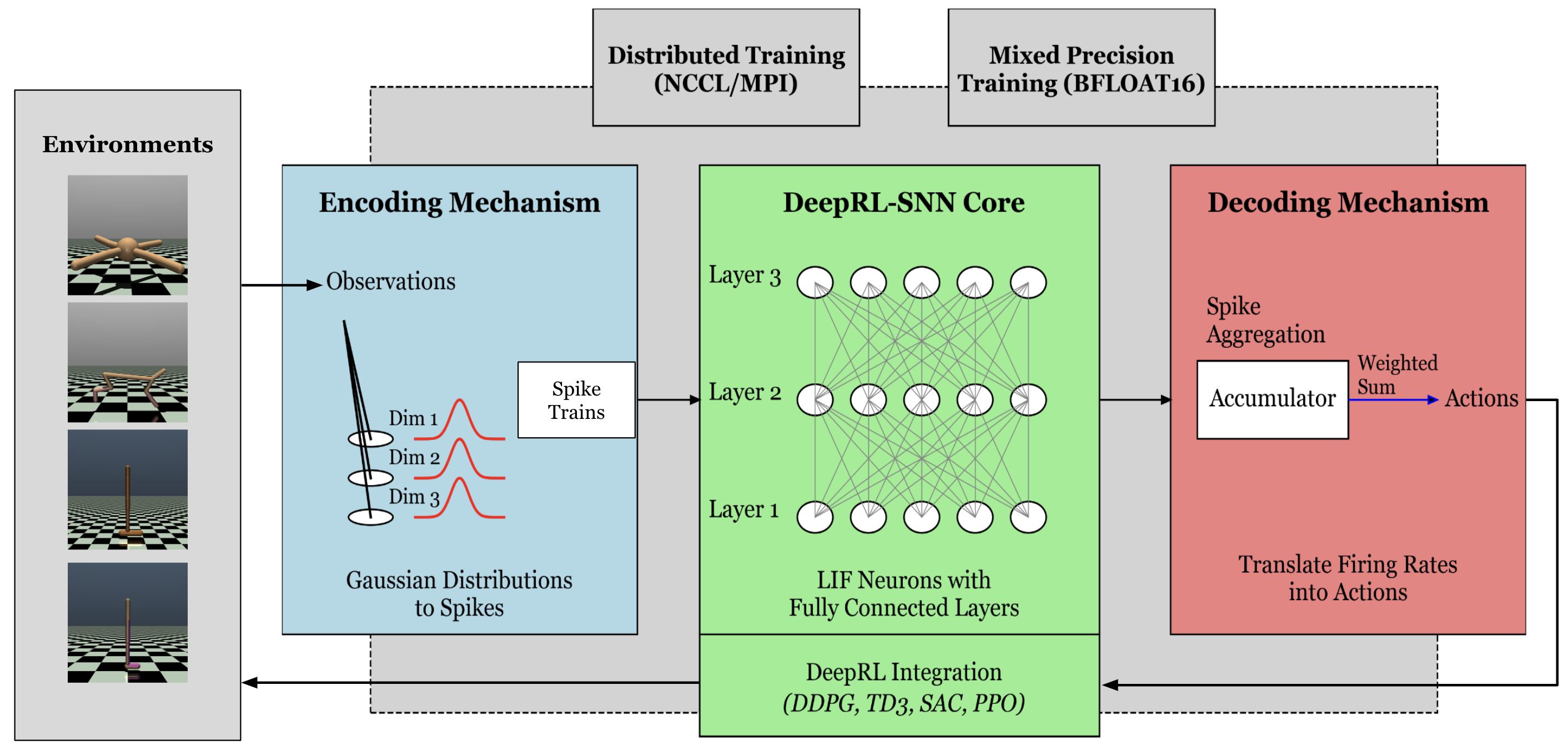}
% \vspace*{-3ex}
\caption{SpikeRL System Architecture.}
% \vspace*{-1ex}
\label{fig:spikeRL}
\end{figure}

The SpikeRL system architecture is illustrated in Figure~\ref{fig:spikeRL}. The SpikeRL framework consists of three major components. First, a DeepRL-based SNN model utilizing population encoding and decoding. Second, distributed training with both Message Passing Interface (MPI) and NVIDIA Collective Communications Library (NCCL) backend is implemented through the PyTorch Distributed package. Lastly, further optimization for model training is achieved by using mixed-precision training with BFLOAT16 data type for parameter updates. 

\subsection{DeepRL-SNN with Population Encoding and Decoding}

The encoding of observations into spikes involves distributing each dimension of the environment’s observation across various neuron populations. These populations use Gaussian distributions to transform continuous input values into spike trains, optimizing the representation capacity and computational efficiency of the network. The core DeepRL-SNN architecture uses a multi-layered, fully-connected SNN with the leaky-integrate-and-fire (LIF)~\cite{b25} model to process the spike trains. This model integrates presynaptic spikes into a current, which then charges the membrane voltage. When the voltage exceeds a predefined threshold, the neuron fires, transmitting a spike to subsequent layers. The decoding process aggregates these spikes over time to compute firing rates, which are then translated into real-valued actions through a weighted sum mechanism. The actions are taken in the environment after being evaluated by the deep critic network in the custom implementation of the DeepRL algorithm, Twin Delayed Deep Deterministic Policy Gradient (TD3)~\cite{b15}.

\section{Distributed Training with MPI and NCCL}

\begin{figure}[tbh]
    \centering
    \includegraphics[width=\linewidth]{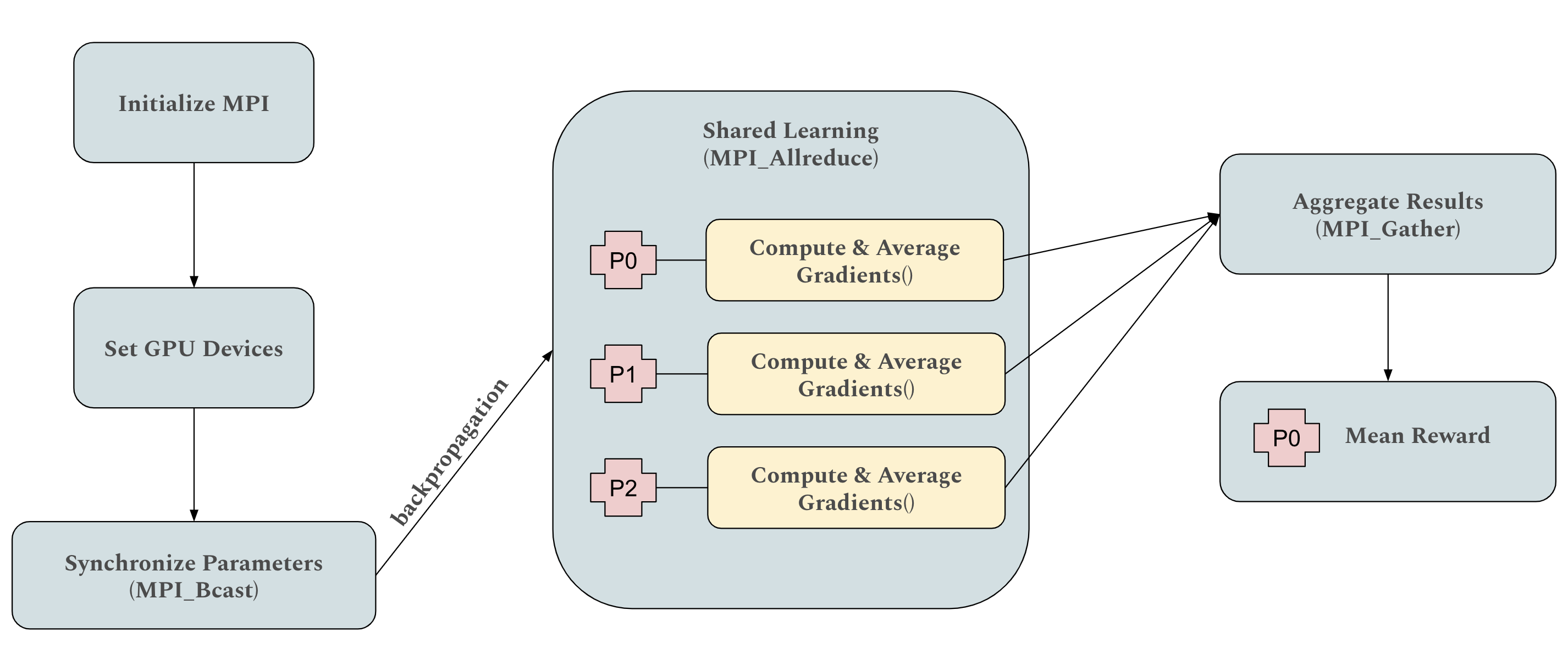}
    \caption{Distributed Training with MPI using \textit{mpi4py}.}
    \label{fig:spikerl_mpi}
\end{figure}

To achieve scalability in training, SpikeRL initially utilized the MPI interface through mpi4py~\cite{b21} as presented in our ICONS24 paper~\cite{b33}, enhancing the training process across multiple computational nodes. In this implementation as shown in Figure~\ref{fig:spikerl_mpi}, at the start of the training, the root process (MPI rank 0) broadcasts the model's parameters to all other processes using the \texttt{broadcast\_weights()} function. This synchronization ensures consistent training conditions across the distributed system. The \texttt{average\_gradients()} function is used to combine gradients from all processes after each has computed gradients based on its data batch. This averaged gradient is then used for model updates, ensuring that all processes contribute equally to the learning process.

\begin{figure}[tbh]
    \centering
    \includegraphics[width=\linewidth]{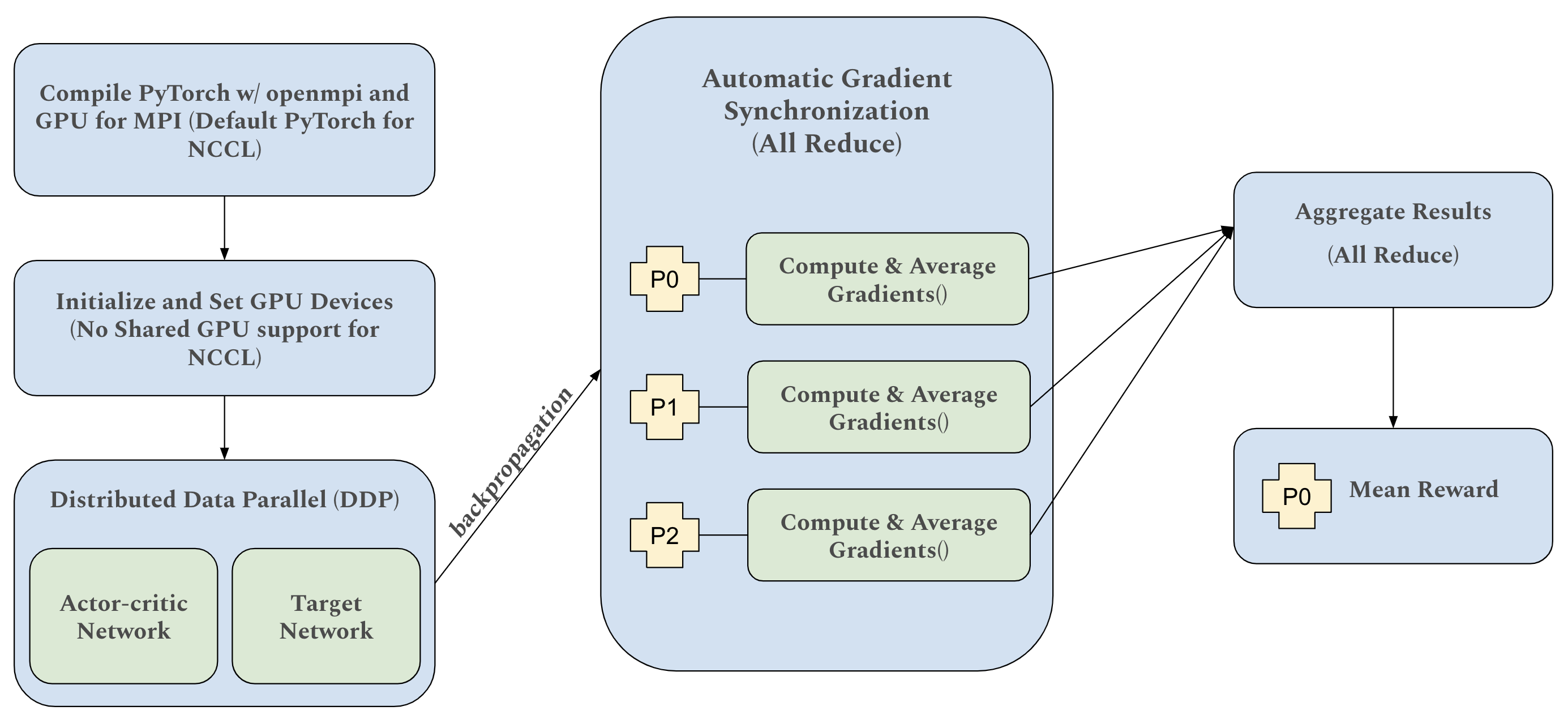}
    \caption{Distributed Training with MPI/NCCL Backend.}
    \label{fig:spikerl_dist}
\end{figure}

Figure~\ref{fig:spikerl_dist} shows a more efficient implementation of distributed training using the distributed package from PyTorch. First, PyTorch was compiled from source with OpenMPI installation for MPI and GPU support and default PyTorch is used for NCCL. After initialization, the “Distributed Data Parallel (DDP)” from torch.distributed is used to wrap both actor-critic and target network model. Then, DDP automatically handles gradient synchronization across processes during backpropagation using the ‘All Reduce’ operation. Using another ‘All Reduce’ operation, the test rewards from all processes are aggregated and the root process calculates the mean reward.

\subsection{Mixed-Precision Optimization with BFLOAT16}

In our implementation, we further refine the efficiency of our
distributed training framework through the incorporation of mixed
precision training techniques. Mixed precision leverages the capability
of modern GPUs to operate on floating-point formats with lower
bit-widths than the standard double or single precision~\cite{b20}. This
approach not only reduces memory usage and bandwidth but also
significantly accelerates mathematical computations, which are central
to the training of deep neural networks~\cite{b20}.

\begin{figure}[tbh]
    \centering
    \includegraphics[width=\linewidth]{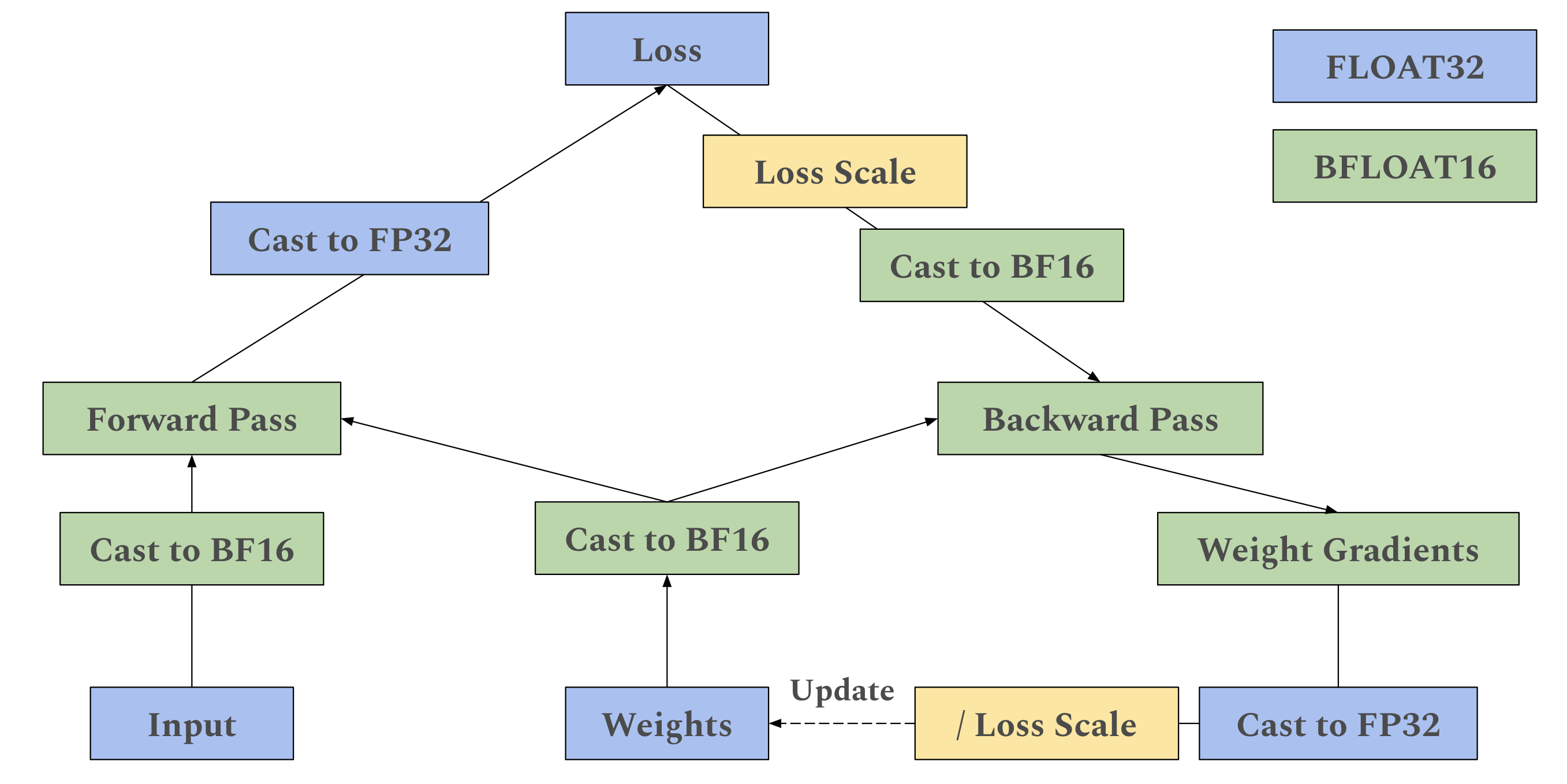}
    \caption{Mixed Precision Training with BFLOAT16.}
    \label{fig:spikerl_mxp}
\end{figure}

We use NVIDIA's Automatic Mixed Precision (AMP) via \textsf{torch.cuda.amp} in our training  as shown in Figure~\ref{fig:spikerl_mxp}, so our model benefits from faster computation with FP16 (half precision) while maintaining the model's accuracy with FP32 (single precision) when necessary. AMP achieves this by maintaining a master copy of weights in full precision, which ensures that weight updates benefit from the full precision's range during the backward pass without losing critical information~\cite{b20}.

The \texttt{autocast} context manager dynamically adjusts the precision during model training, allowing for efficient execution of operations in the appropriate precision level.
To address the reduced precision's potential issues, such as gradient underflow, the \texttt{GradScaler} is utilized. This tool scales the loss function during the backward pass to maintain gradient magnitude, adjusting the scale as necessary to prevent overflow.

\section{Experimental Setup}

\subsection{Energy Measurements}
We utilized the CodeCarbon tool to measure total energy consumption using data from Intel Running Average Power Limit (RAPL) NVIDIA Management Library (NVML). Then, we used the Greenup energy efficiency metric which is calculated as,
\[
\text{Greenup} = \frac{\text{Speedup}\!\left(\frac{T_\phi}{T_0}\right)}
                   {\text{Powerup}\!\left(\frac{P_0}{P_\phi}\right)}
\]

Using the Greenup metric we can plot the Greenup, Powerup and Speedup (GPS-UP) quadrant graph that shows four energy saving zones as green zones with four red zones that waste energy.

\subsection{Environment Configuration}
For the performance and energy efficiency benchmarks, ICL's Guyot cluster with 8 Nvidia A100 80GB GPUs were used. For our continuous control experiments the following gymnasium~\cite{b26} Mujoco~\cite{b22} environments were chosen:

\textbf{Ant-v4:} A 3D quadruped robot with 8 hinges. It can take actions in the environment by applying torque to it's 8 hinges. It has 105 observation space with position, velocity, and mass values.

\textbf{HalfCheetah-v4:} A 2D robot with 6 hinges. It can take actions in the environment by applying torque to it's 6 hinges to move forward as fast as possible. It has 17 observation space with position and velocity values.

\textbf{Hopper-v4:} A 2D one-legged robot with 3 hinges. It can take actions in the environment to hop forward by applying torque to it's 3 hinges. It has 11 observation space with position and velocity values.

\textbf{Walker2D-v4:} A 2D two-legged robot with 6 hinges. It can take actions in the environment by applying torque to it's 6 hinges to move forward. It has 17 observation space with position and velocity values.

The reward schedule encourages this behavior: positive rewards for forward movement and survival at each timestep, and negative rewards for excessively large, destabilizing actions. Good performance is characterized by stable, consistent forward movement and efficient action optimization to maximize cumulative reward. These continuous control problems serve as standard benchmarks within the reinforcement learning community.

\section{Results}

\subsection{Performance Benchmarks}
\begin{figure}[tbh]
    \centering
    \includegraphics[width=\linewidth]{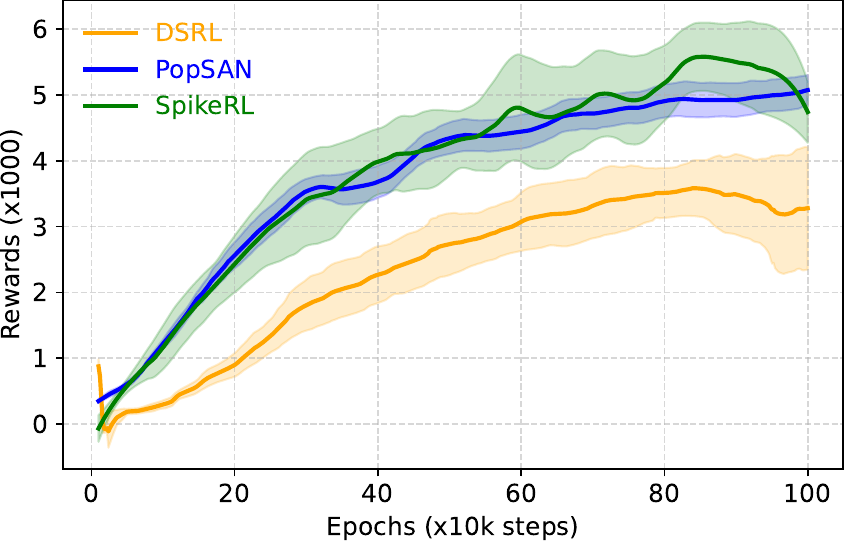}
    \caption{Ant-v4 Performance Benchmark of SpikeRL over PopSAN and DSRL.}
    \label{fig:ant}
\end{figure}

\begin{figure}[tbh]
    \centering
    \includegraphics[width=\linewidth]{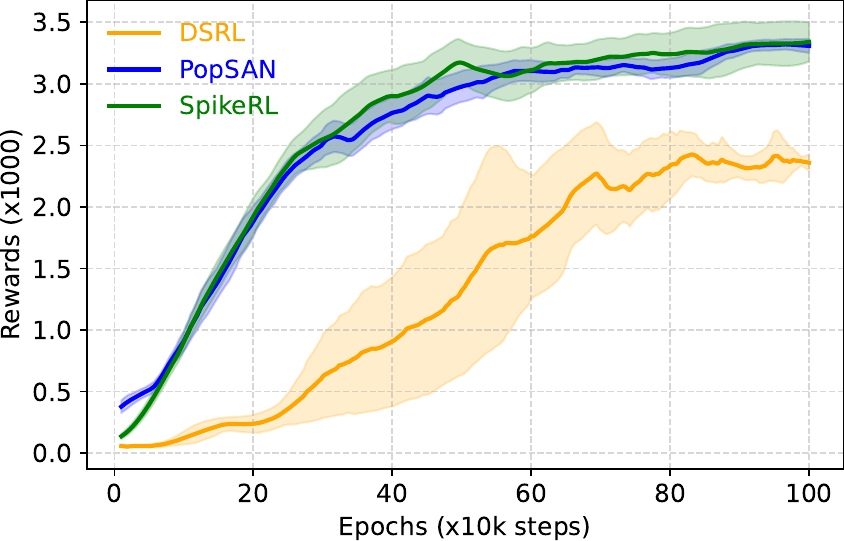}
    \caption{Hopper-v4 Performance Benchmark of SpikeRL over PopSAN and DSRL.}
    \label{fig:hopper}
\end{figure}

\begin{figure}[tbh]
    \centering
    \includegraphics[width=\linewidth]{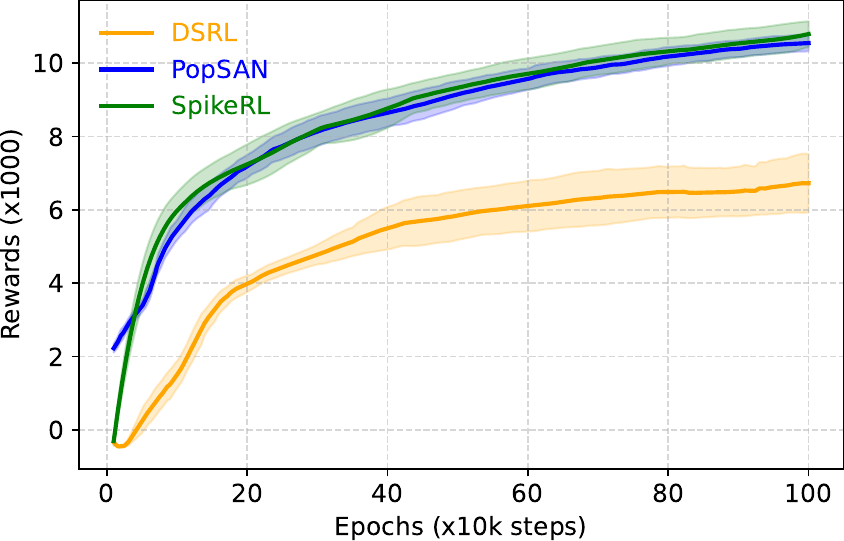}
    \caption{HalfCheetah-v4 Performance Benchmark of SpikeRL over PopSAN and DSRL.}
    \label{fig:halfcheetah}
\end{figure}

Figures~\ref{fig:ant},~\ref{fig:hopper}, and~\ref{fig:halfcheetah} shows SpikeRL performance benchmark of 5 runs for the Ant, Hopper, and HalfCheetah environment over PopSAN and DSRL methods. For Ant environment, SpikeRL achieves 3.14\% higher mean reward compared to PopSAN and achieves 65.62\% higher mean reward compared to DSRL. In Hopper environment, SpikeRL achieves 2.01\% higher mean reward compared to PopSAN and achieves 98.26\% higher mean reward compared to DSRL. For HalfCheetah, SpikeRL achieves 1.51\% higher mean reward compared to PopSAN and achieves 70.91\% higher mean reward compared to DSRL.

\subsection{Energy Efficiency Benchmarks}
The results of the energy efficiency benchmarks for SpikeRL (NCCL) and SpikeRL (MPI) over PopSAN in the OpenAI Gym environments, Ant-v4, Hopper-v4, and HalfCheetah-v4 are highlighted in Tables~\ref{tab:Ant},~\ref{tab:Hopper}, and~\ref{tab:HalfCheetah} respectively. Each benchmark includes metrics for execution time, energy consumption, and average power consumption, with further GPS-UP metric calculations for speedup, powerup, and greenup ratios to compare the frameworks.

For the Ant-v4 environment, SpikeRL (NCCL) demonstrates a substantial improvement over PopSAN, achieving a speedup of 1.99×. Though the powerup also goes up to 1.53× suggesting more power consumption, the trade-off results in a  greenup of 1.30×. SpikeRL (MPI) also shows efficiency gains over PopSAN with a speedup of 1.84×. With a powerup of 1.49× it was also consuming more power than PopSAN, ultimately resulting in a greenup of 1.24×. When comparing SpikeRL (NCCL) to SpikeRL (MPI), the NCCL version exhibits a modest advantage with a speedup of 1.08×. However, it also gained a powerup of 1.03× demonstrating more power consumption than the MPI model. Overall, with a greenup of 1.05×, the NCCL model was slightly more energy efficient than the MPI model.

\begin{table}[tbh]
\centering
\caption{Energy Efficiency Benchmark for OpenAI Gym Ant-v4}
\label{tab:Ant}
\begin{tabularx}{\linewidth}{p{3.8cm} >{\raggedleft\arraybackslash}X >{\raggedleft\arraybackslash}X >{\raggedleft\arraybackslash}X}
\hline
\textbf{Metric} & \textbf{PopSAN} & \textbf{SpikeRL (NCCL)} & \textbf{SpikeRL (MPI)} \\
\hline
Time (hours)                 & 11.92 & 6.00  & 6.46 \\
Energy Consumed (kWh)        & 18.70 & 14.40 & 15.11 \\
Avg Power Consumption (kW)   & 1.57  & 2.40  & 2.34 \\
\hline
\multicolumn{4}{l}{\textbf{GPS-UP Metric Calculations}} \\
\hline
\textbf{Comparison} & \textbf{Speedup} & \textbf{Powerup} & \textbf{Greenup} \\
\hline
SpikeRL (NCCL) over PopSAN          & $1.99 \times$ & $1.53 \times$ & $1.30 \times$ \\
SpikeRL (MPI) over PopSAN           & $1.84 \times$ & $1.49 \times$ & $1.24 \times$ \\
SpikeRL (NCCL over MPI)   & $1.08 \times$ & $1.03 \times$ & $1.05 \times$ \\
\hline
\end{tabularx}
\end{table}

In the Hopper-v4 environment, SpikeRL (NCCL) improves over PopSAN with a speedup of 1.30×. In this case, it also consumes less power than PopSAN with it's powerup being slightly below 1 at 0.93×. The combination of this results in a higher greenup than the Ant-v4 environment with it's value reaching at 1.39×. SpikeRL (MPI) displays a slight edge in performance with a speedup of 1.37×, a powerup of 0.97×, and a greenup of 1.41×. Comparing the two SpikeRL methods, the MPI model had a marginal energy efficiency gain over the NCCL model.

\begin{table}[tbh]
\centering
\caption{Energy Efficiency Benchmark for OpenAI Gym Hopper-v4}
\label{tab:Hopper}
\begin{tabularx}{\linewidth}{p{3.8cm} >{\raggedleft\arraybackslash}X >{\raggedleft\arraybackslash}X >{\raggedleft\arraybackslash}X}
\hline
\textbf{Metric} & \textbf{PopSAN} & \textbf{SpikeRL (NCCL)} & \textbf{SpikeRL (MPI)} \\
\hline
Time (hours)           & 8.52 & 6.57 & 6.22 \\
Energy Consumed (kWh)      & 18.93 & 13.63 & 13.45 \\
Avg Power Consumption (kW)   & 2.22  & 2.07 & 2.16 \\
\hline
\multicolumn{4}{l}{\textbf{GPS-UP Metric Calculations}} \\
\hline
\textbf{Comparison} & \textbf{Speedup} & \textbf{Powerup} & \textbf{Greenup} \\
\hline
SpikeRL (NCCL) over PopSAN   & $1.30 \times$ & $0.93 \times$ & $1.39 \times$ \\
SpikeRL (MPI) over PopSAN    & $1.37 \times$ & $0.97 \times$ & $1.41 \times$ \\
SpikeRL (NCCL over MPI)  & $0.95 \times$ & $0.96 \times$ & $0.99 \times$ \\
\hline
\end{tabularx}
\end{table}

For HalfCheetah-v4, SpikeRL (NCCL) demonstrates a speedup of 1.48×, a powerup of 0.98× and a greenup of 1.52× over PopSAN. SpikeRL (MPI) shows similar gains over PopSAN, with a speedup of 1.41×, powerup of 0.96×, and greenup of 1.47×. Between the two SpikeRL models, NCCL is slightly more energy efficient in this case, with a speedup of 1.05×, powerup of 1.01×, and greenup of 1.03×.

\begin{table}[tbh]
\centering
\caption{Energy Efficiency Benchmark for OpenAI Gym HalfCheetah-v4}
\label{tab:HalfCheetah}
\begin{tabularx}{\linewidth}{p{3.8cm} >{\raggedleft\arraybackslash}X >{\raggedleft\arraybackslash}X >{\raggedleft\arraybackslash}X}
\hline
\textbf{Metric} & \textbf{PopSAN} & \textbf{SpikeRL (NCCL)} & \textbf{SpikeRL (MPI)} \\
\hline
Time (hours)           & 9.09 & 6.14 & 6.43 \\
Energy Consumed (kWh)      & 19.73 & 12.99 & 13.44 \\
Avg Power Consumption (kW)   & 2.17  & 2.12 & 2.09 \\
\hline
\multicolumn{4}{l}{\textbf{GPS-UP Metric Calculations}} \\
\hline
\textbf{Comparison} & \textbf{Speedup} & \textbf{Powerup} & \textbf{Greenup} \\
\hline
SpikeRL (NCCL) over PopSAN   & $1.48 \times$ & $0.98 \times$ & $1.52 \times$ \\
SpikeRL (MPI) over PopSAN    & $1.41 \times$ & $0.96 \times$ & $1.47 \times$ \\
SpikeRL (NCCL over MPI)  & $1.05 \times$ & $1.01 \times$ & $1.03 \times$ \\
\hline
\end{tabularx}
\end{table}

\begin{figure}[tbh]
    \centering
    \includegraphics[width=\linewidth]{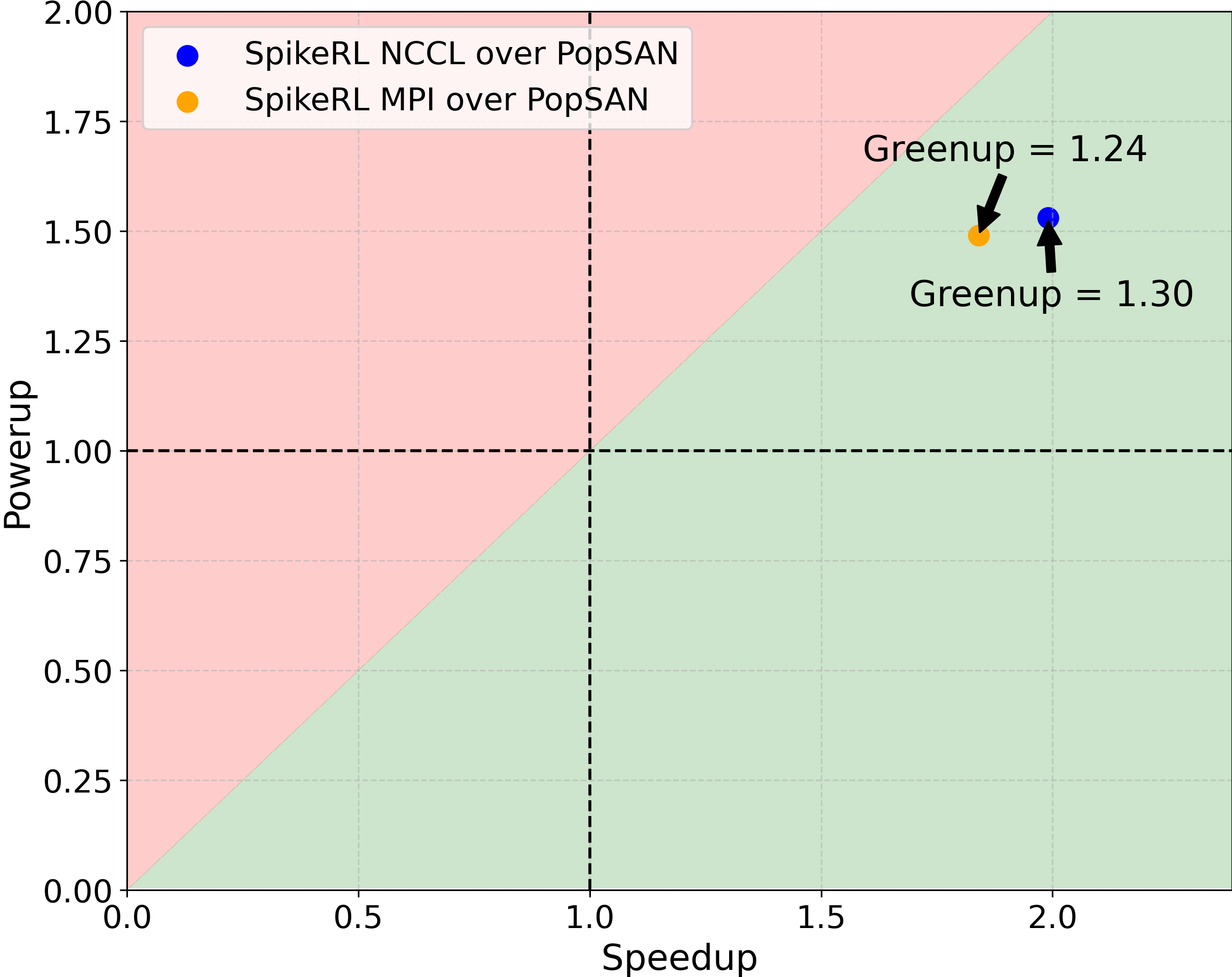}
    \caption{SpikeRL Ant-v4 GPS-UP Quadrant Graph.}
    \label{fig:ant_gps}
\end{figure}

The GPS-UP quadrant graphs visually represent the greenup metric with powerup versus speedup across the three environments Ant-v4, Hopper-v4, and HalfCheetah-v4. 
In Ant-v4 GPS-UP quadrant graph in Figure~\ref{fig:ant_gps}, SpikeRL (NCCL) and SpikeRL (MPI) are positioned in the green zone at the top right quadrant indicating that the energy efficiency is largely being attributed to their speedup over PopSAN rather than their powerup. 

\begin{figure}[tbh]
    \centering
    \includegraphics[width=\linewidth]{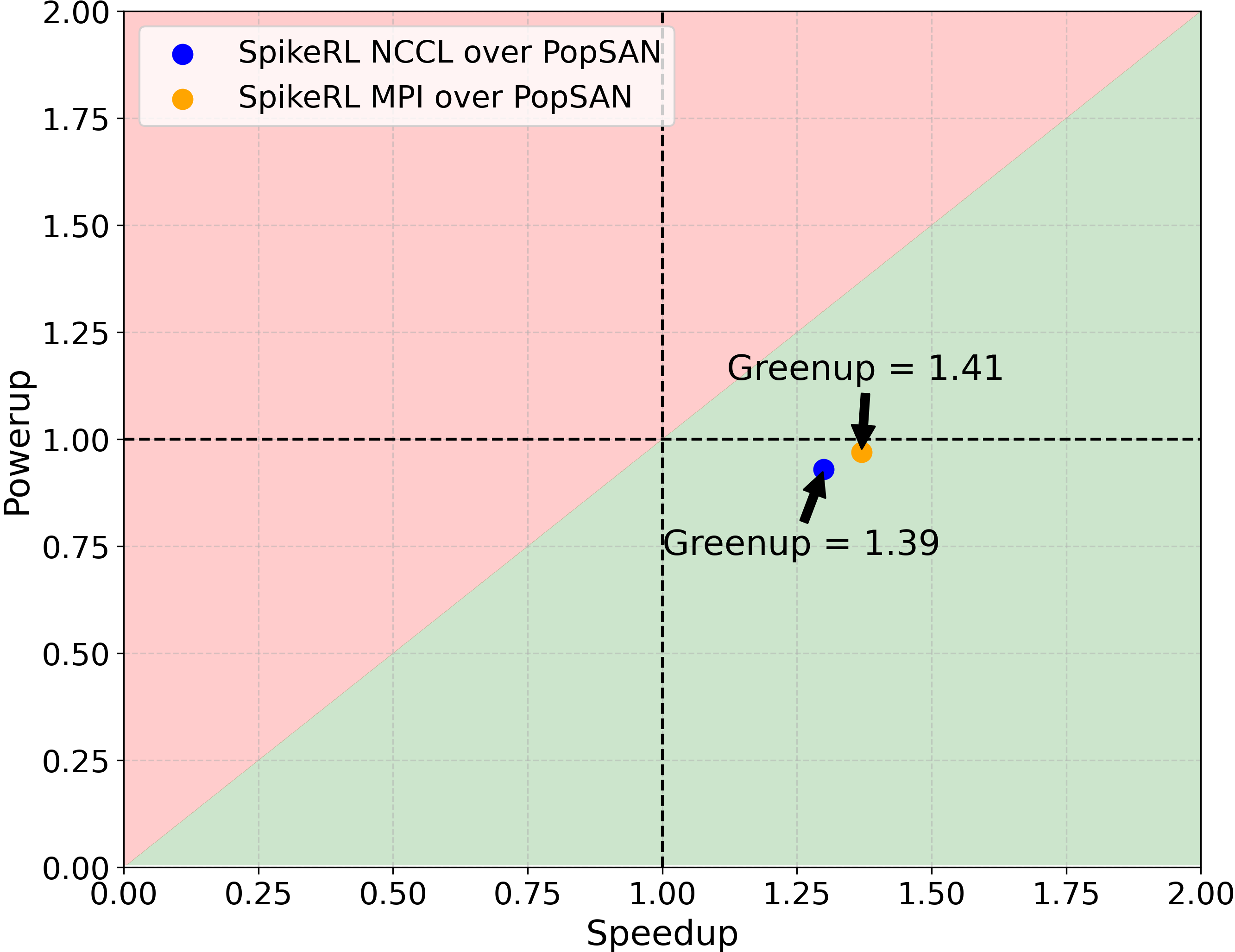}
    \caption{SpikeRL Hopper-v4 GPS-UP Quadrant Graph.}
    \label{fig:hopper_gps}
\end{figure}

In the case of Hopper-v4 GPS-UP quadrant graph in Figure~\ref{fig:hopper_gps}, the greenup of both MPI and NCCL are in the all green zone of bottom right quadrant. This zone signifies that the energy efficiency in this case is a result of both speedup gain and from less power consumption. 

\begin{figure}[tbh]
    \centering
    \includegraphics[width=\linewidth]{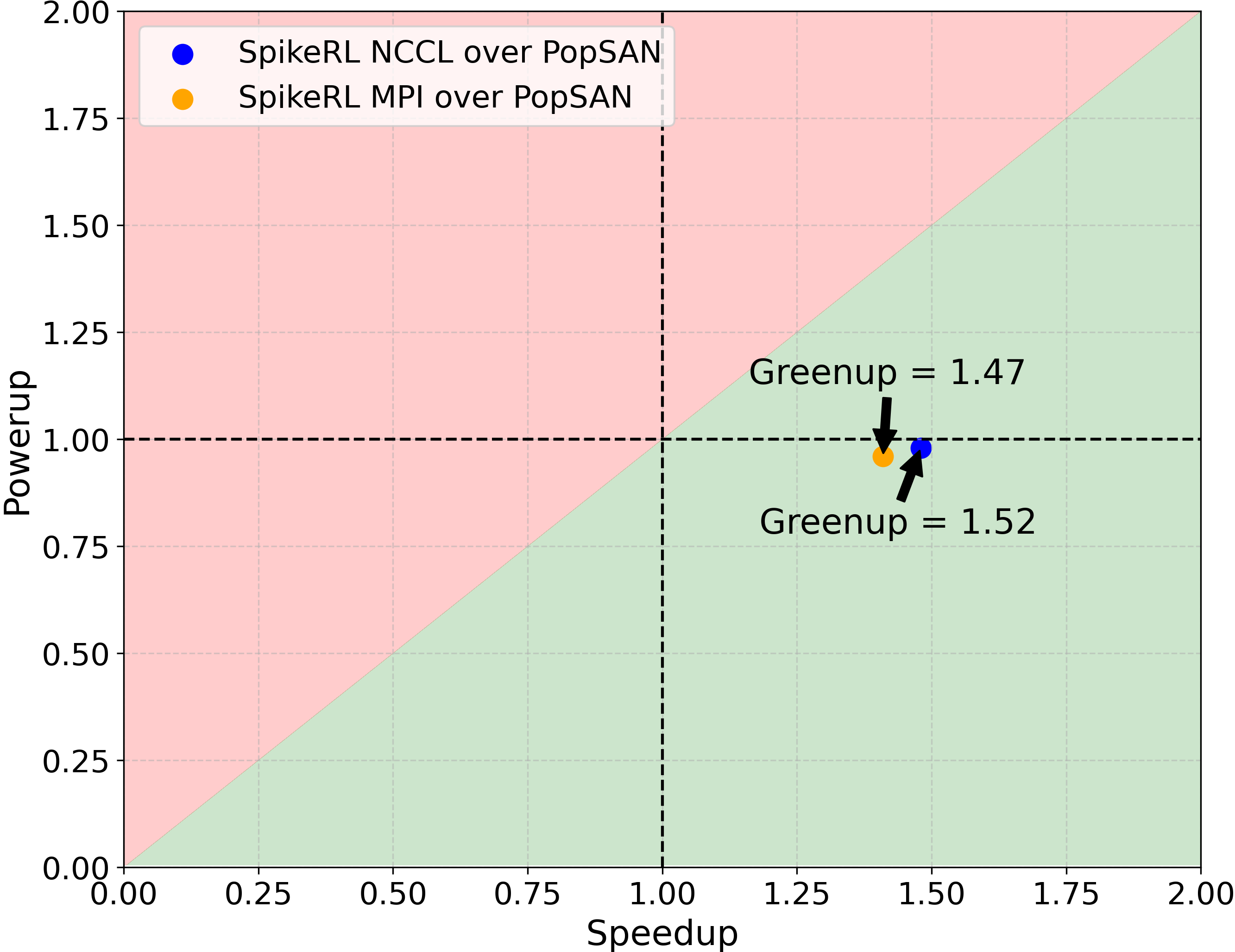}
    \caption{SpikeRL HalfCheetah-v4 GPS-UP Quadrant Graph.}
    \label{fig:halfcheetah_gps}
\end{figure}

Figure~\ref{fig:halfcheetah_gps} shows the GPS-UP quadrant graph of HalfCheetah-v4, where similarly to the Hopper-v4 environment, both SpikeRL models are in the all green zone with an even higher greenup value indicating more energy efficiency over PopSAN compared to the case of Ant-v4 and Hopper-v4 environments.

These energy efficiency gains are reflected in the carbon footprint analysis which highlights the environmental benefits of using SpikeRL over PopSAN across these environments. 

\begin{figure}[tbh]
    \centering
    \includegraphics[width=\linewidth]{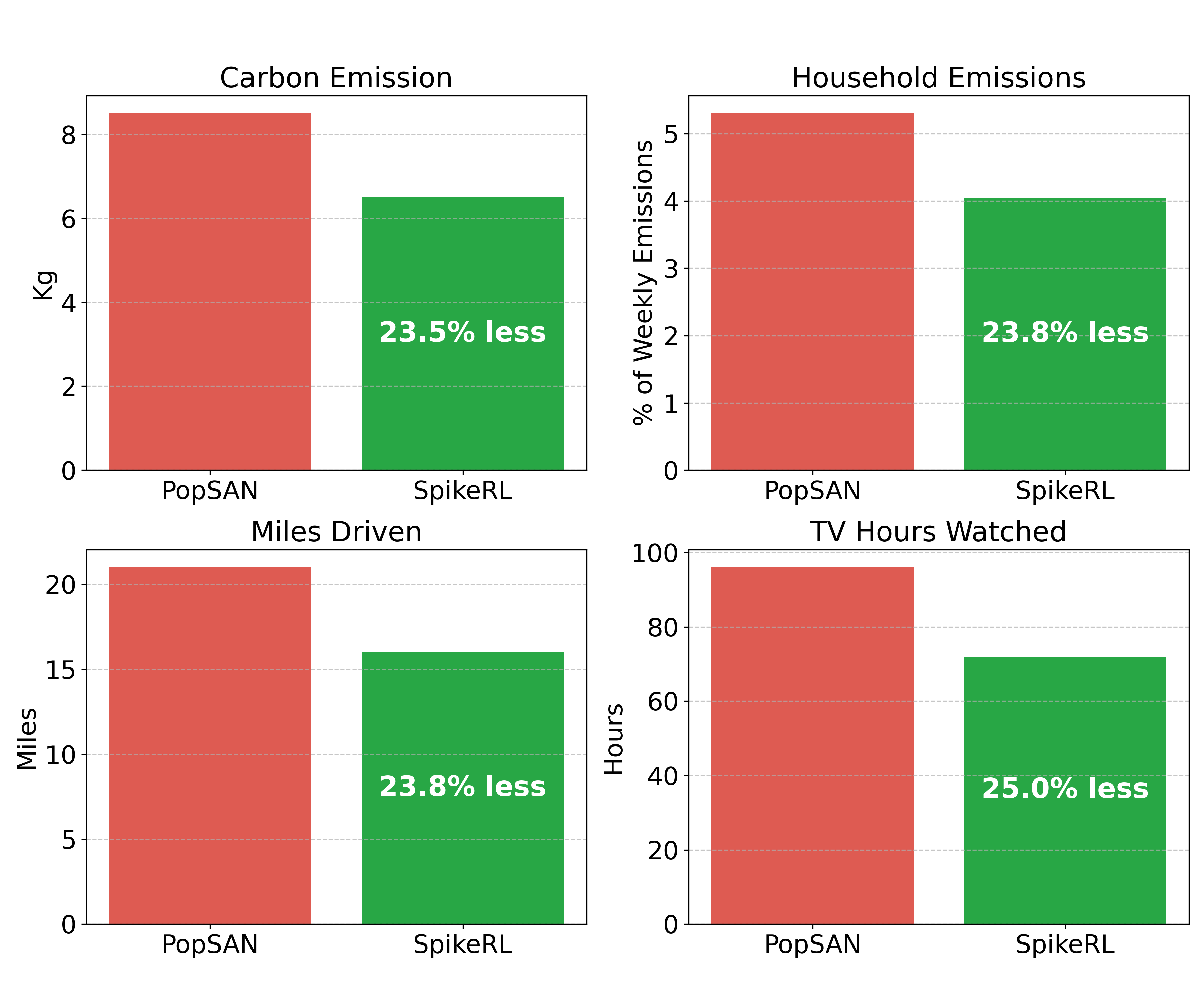}
    \caption{Ant-v4 Carbon Footprint Analysis of SpikeRL over PopSAN.}
    \label{fig:ant_carbon}
\end{figure}

Figure~\ref{fig:ant_carbon} demonstrates carbon footprint analysis in the Ant-v4 environment where where SpikeRL reduces carbon emission by 23.5\% compared to PopSAN which is equivalent to 23.8\% less weekly household emissions, 23.8\% less miles driven of a gasoline car, and 25\% less tv hours watched. 

\begin{figure}[tbh]
    \centering
    \includegraphics[width=\linewidth]{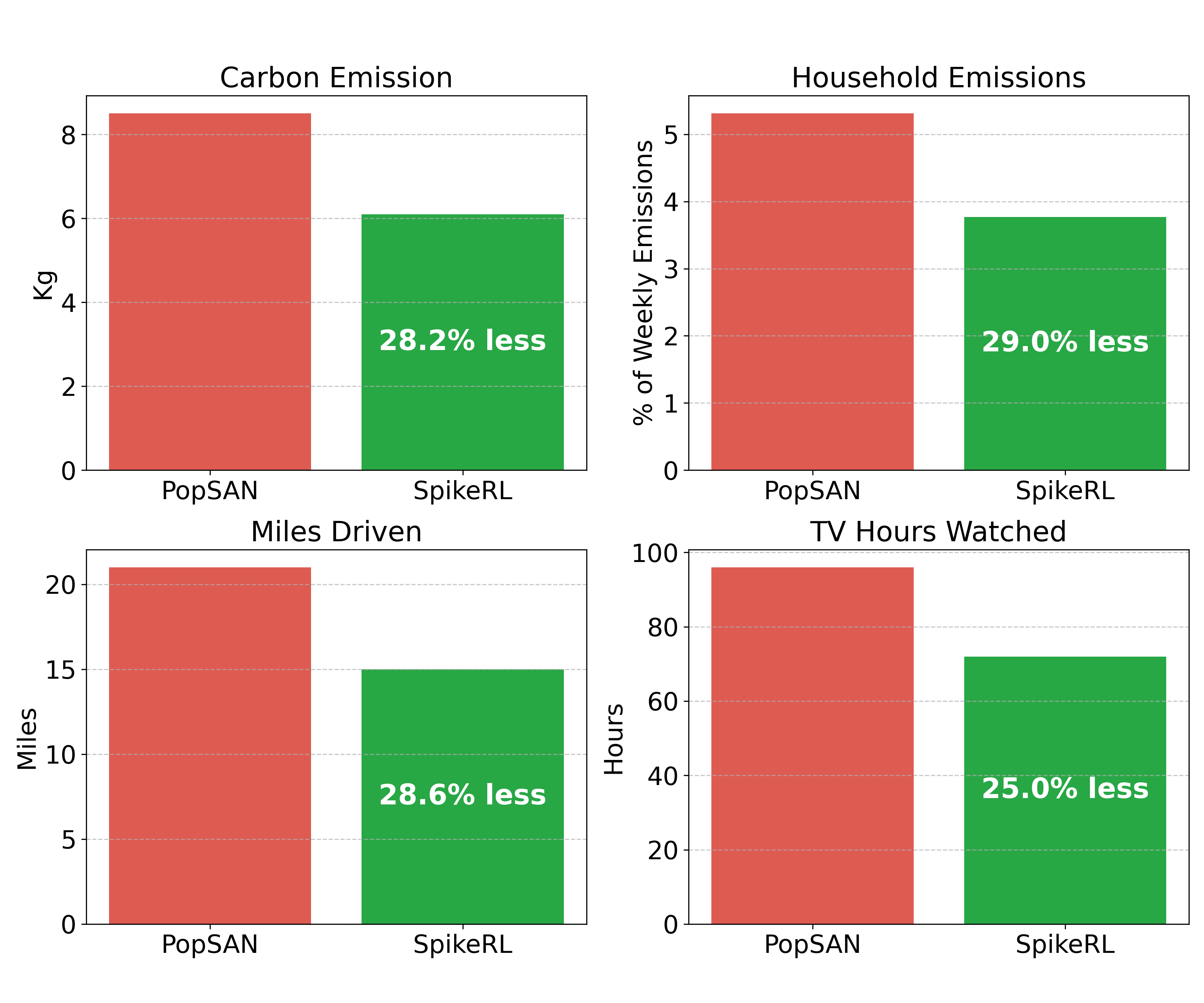}
    \caption{Hopper-v4 Carbon Footprint Analysis of SpikeRL over PopSAN.}
    \label{fig:hopper_carbon}
\end{figure}

In Hopper-v4 in Figure~\ref{fig:hopper_carbon}, the reductions of SpikeRL are slightly more pronounced, with decreases around 28-29\% for carbon and household emissions. 

\begin{figure}[tbh]
    \centering
    \includegraphics[width=\linewidth]{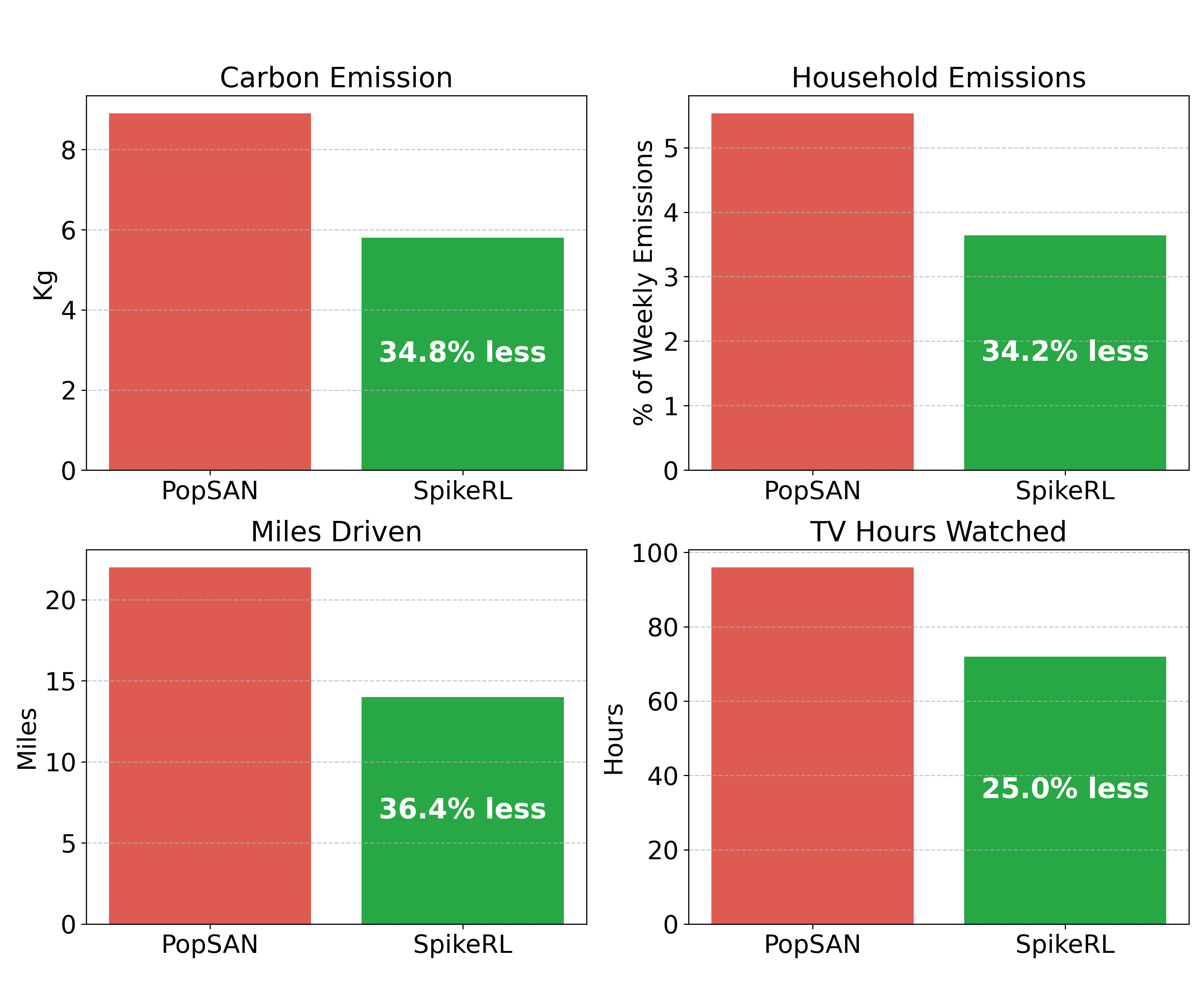}
    \caption{HalfCheetah-v4 Carbon Footprint Analysis of SpikeRL over PopSAN.}
    \label{fig:halfcheetah_carbon}
\end{figure}

HalfCheetah-v4 in Figure~\ref{fig:halfcheetah_carbon} showcases the most significant reductions, with a 34.8\% decrease in carbon emission, 34.2\% decrease in weekly household emissions, 36.4\% less miles driven, and a similar 25\% TV hours watched. 

These results illustrate the substantial environmental impact of using SpikeRL over PopSAN. These benchmarks underscore the energy efficiency and environmental advantages of the SpikeRL framework, for training continuous control agents across diverse OpenAI Gym environments.

\section{Conclusion}

In terms of performance benchmarks, the results presented demonstrate a significant amount of performance boost over the DSRL model with an average of 78.23\% more performance. Comparison with the PopSAN method showed only a moderate amount of performance improvement with an average of 2.22\% improved performance. This similarity in performance to PopSAN can be attributed to the population encoding and decoding mechanism which is present in both SpikeRL and PopSAN.

Although SpikeRL performed somewhat similarly to PopSAN in accumulating rewards, energy efficiency benchmarks demonstrated a superior energy efficiency of SpikeRL over PopSAN. With an average of 1.39x more energy efficiency than PopSAN, SpikeRL managed to reduce the carbon emissions by up to 34.8\%. Thus SpikeRL proved to be an energy efficient solution with losing scalability or performance.

In future work, we aim to get the best performance results from SpikeRL training by extensive hyperparameter tuning to ensure enhanced scalability. Another goal is to test SpikeRL on more complex continuous control environments such as the Humanoid environment from Mujoco. Lastly, run inference of SpikeRL model on a simulated neuromorphic processor and finally deploy on a neuromorphic hardware.

% \IEEEtriggeratref{12}

\bibliographystyle{IEEEtran}
\bibliography{nice25spikerl}

\begin{thebibliography}{10}
\providecommand{\url}[1]{#1}
\csname url@samestyle\endcsname
\providecommand{\newblock}{\relax}
\providecommand{\bibinfo}[2]{#2}
\providecommand{\BIBentrySTDinterwordspacing}{\spaceskip=0pt\relax}
\providecommand{\BIBentryALTinterwordstretchfactor}{4}
\providecommand{\BIBentryALTinterwordspacing}{\spaceskip=\fontdimen2\font plus
\BIBentryALTinterwordstretchfactor\fontdimen3\font minus \fontdimen4\font\relax}
\providecommand{\BIBforeignlanguage}[2]{{%
\expandafter\ifx\csname l@#1\endcsname\relax
\typeout{** WARNING: IEEEtran.bst: No hyphenation pattern has been}%
\typeout{** loaded for the language `#1'. Using the pattern for}%
\typeout{** the default language instead.}%
\else
\language=\csname l@#1\endcsname
\fi
#2}}
\providecommand{\BIBdecl}{\relax}
\BIBdecl

\bibitem{b27}
Y.~I. Alzoubi and A.~Mishra, ``Green artificial intelligence initiatives: Potentials and challenges,'' \emph{Journal of Cleaner Production}, p. 143090, 2024.

\bibitem{b34}
W.~{da Silva Pereira}, J.~Silva, and T.~Tahmid, ``\BIBforeignlanguage{American English}{Low precision for lower energy consumption: Preprint},'' in \emph{\BIBforeignlanguage{American English}{Low Precision for Lower Energy Consumption: Preprint}}, 2024, aSCR Workshop on Energy-Efficient Computing for Science ; Conference date: 09-09-2024 Through 12-09-2024.

\bibitem{b28}
C.~D. Schuman, T.~E. Potok, R.~M. Patton, J.~D. Birdwell, M.~E. Dean, G.~S. Rose, and J.~S. Plank, ``A survey of neuromorphic computing and neural networks in hardware,'' \emph{arXiv preprint arXiv:1705.06963}, 2017.

\bibitem{b29}
C.~D. Schuman, S.~R. Kulkarni, M.~Parsa, J.~P. Mitchell, B.~Kay \emph{et~al.}, ``Opportunities for neuromorphic computing algorithms and applications,'' \emph{Nature Computational Science}, vol.~2, no.~1, pp. 10--19, 2022.

\bibitem{b30}
A.~P. Georgopoulos, A.~B. Schwartz, and R.~E. Kettner, ``Neuronal population coding of movement direction,'' \emph{Science}, vol. 233, no. 4771, pp. 1416--1419, 1986.

\bibitem{b31}
D.~W. Walker and J.~J. Dongarra, ``Mpi: a standard message passing interface,'' \emph{Supercomputer}, vol.~12, pp. 56--68, 1996.

\bibitem{b20}
P.~Micikevicius, S.~Narang, J.~Alben, G.~Diamos, E.~Elsen, D.~Garcia, and H.~Wu, ``Mixed precision training,'' \emph{arXiv preprint}, vol. 1710.03740, 2017.

\bibitem{b32}
C.~D. Schuman, S.~R. Kulkarni, M.~Parsa, J.~P. Mitchell, B.~Kay \emph{et~al.}, ``Opportunities for neuromorphic computing algorithms and applications,'' \emph{Nature Computational Science}, vol.~2, no.~1, pp. 10--19, 2022.

\bibitem{b1}
K.~Yamazaki, V.~K. Vo-Ho, D.~Bulsara, and N.~Le, ``Spiking neural networks and their applications: A review,'' \emph{Brain Sciences}, vol.~12, no.~7, p. 863, 2022.

\bibitem{b2}
B.~Cuevas-Arteaga, J.~P. Dominguez-Morales, H.~Rostro-Gonzalez, A.~Espinal, A.~F. Jimenez-Fernandez, F.~Gomez-Rodriguez, and A.~Linares-Barranco, ``A {SpiNNaker} application: design, implementation and validation of {SCPGs},'' in \emph{In Advances in Computational Intelligence: 14th International Work-Conference on Artificial Neural Networks, IWANN 2017, Proceedings, Part I}, vol.~14.\hskip 1em plus 0.5em minus 0.4em\relax Cadiz, Spain: Springer International Publishing, June 14-16 2017, pp. 548--559.

\bibitem{b3}
B.~Strohmer, P.~Manoonpong, and L.~B. Larsen, ``Flexible spiking {CPGs} for online manipulation during hexapod walking,'' \emph{Frontiers in Neurorobotics}, vol.~14, no.~41, 2020.

\bibitem{b4}
S.~B. Furber, F.~Galluppi, S.~Temple, and L.~A. Plana, ``The {SpiNNaker} project,'' \emph{Proceedings of the IEEE}, vol. 102, no.~5, pp. 652--665, 2014.

\bibitem{b5}
E.~Donati, F.~Corradi, C.~Stefanini, and G.~Indiveri, ``A spiking implementation of the lamprey's central pattern generator in neuromorphic {VLSI},'' in \emph{2014 IEEE Biomedical Circuits and Systems Conference (BioCAS) Proceedings}.\hskip 1em plus 0.5em minus 0.4em\relax IEEE, October 2014, pp. 512--515.

\bibitem{b6}
E.~Angelidis, E.~Buchholz, J.~Arreguit, A.~Rougé, T.~Stewart, A.~von Arnim, and A.~Ijspeert, ``A spiking central pattern generator for the control of a simulated lamprey robot running on {SpiNNaker} and {Loihi} neuromorphic boards,'' \emph{Neuromorphic Computing and Engineering}, vol.~1, no.~1, p. 014005, 2021.

\bibitem{b7}
J.~Dupeyroux, J.~J. Hagenaars, F.~Paredes-Vallés, and G.~C. de~Croon, ``Neuromorphic control for optic-flow-based landing of {MAVs} using the {Loihi} processor,'' in \emph{2021 IEEE International Conference on Robotics and Automation (ICRA)}.\hskip 1em plus 0.5em minus 0.4em\relax IEEE, May 2021, pp. 96--102.

\bibitem{b8}
R.~K. Stagsted, A.~Vitale, A.~Renner, L.~B. Larsen, A.~L. Christensen, and Y.~Sandamirskaya, ``Event-based {PID} controller fully realized in neuromorphic hardware: A one {DoF} study,'' in \emph{2020 IEEE/RSJ international conference on intelligent robots and systems (IROS)}.\hskip 1em plus 0.5em minus 0.4em\relax IEEE, October 2020, pp. 10\,939--10\,944.

\bibitem{b9}
F.~Galluppi, J.~Conradt, T.~Stewart, C.~Eliasmith, T.~Horiuchi, J.~Tapson, and R.~Etienne-Cummings, ``Live demo: Spiking {ratSLAM}: Rat hippocampus cells in spiking neural hardware,'' in \emph{2012 IEEE Biomedical Circuits and Systems Conference (BioCAS)}.\hskip 1em plus 0.5em minus 0.4em\relax IEEE, November 2012, pp. 91--91.

\bibitem{b10}
G.~Tang and K.~P. Michmizos, ``Gridbot: An autonomous robot controlled by a spiking neural network mimicking the brain's navigational system,'' in \emph{Proceedings of the International Conference on Neuromorphic Systems}, July 2018, pp. 1--8.

\bibitem{b11}
Z.~Bing, I.~Baumann, Z.~Jiang, K.~Huang, C.~Cai, and A.~Knoll, ``Supervised learning in snn via reward-modulated spike-timing-dependent plasticity for a target reaching vehicle,'' \emph{Frontiers in Neurorobotics}, vol.~13, no.~18, 2019.

\bibitem{b12}
G.~Tang, N.~Kumar, and K.~P. Michmizos, ``Reinforcement co-learning of deep and spiking neural networks for energy-efficient mapless navigation with neuromorphic hardware,'' \emph{arXiv preprint}, vol. 2003.01157, 2020.

\bibitem{b13}
G.~Tang, N.~Kumar, R.~Yoo, and K.~Michmizos, ``Deep reinforcement learning with population-coded spiking neural network for continuous control,'' in \emph{Conference on Robot Learning}, October 2021, pp. 2016--2029, pMLR.

\bibitem{b18}
M.~Davies, Srinivasa, L.~N., C.~T.~H., C.~G., C.~Y., S.~H., and H.~Wang, ``Loihi: A neuromorphic manycore processor with on-chip learning,'' \emph{IEEE Micro}, vol.~38, no.~1, pp. 82--99, 2018.

\bibitem{b35}
J.~V. de~Oliveira~Silva, T.~Tahmid, and W.~da~Silva~Pereira, ``Low precision and efficient programming languages for sustainable ai: Final report for the summer project of 2024,'' National Renewable Energy Laboratory (NREL), Golden, CO (United States), Tech. Rep., 2024.

\bibitem{b25}
M.~A. Nahmias, B.~J. Shastri, A.~N. Tait, and P.~R. Prucnal, ``A leaky integrate-and-fire laser neuron for ultrafast cognitive computing,'' \emph{IEEE journal of selected topics in quantum electronics}, vol.~19, no.~5, pp. 1--12, 2013.

\bibitem{b15}
S.~Fujimoto, H.~Hoof, and D.~Meger, ``Addressing function approximation error in actor-critic methods,'' in \emph{International conference on machine learning}, July 2018, pp. 1587--1596, pMLR.

\bibitem{b21}
L.~Dalcin and Y.~L.~L. Fang, ``mpi4py: Status update after 12 years of development,'' \emph{Computing in Science \& Engineering}, vol.~23, no.~4, pp. 47--54, 2021.

\bibitem{b33}
T.~{Tahmid}, M.~Gates, P.~Luszczek, and C.~D. Schuman, ``Towards scalable and efficient spiking reinforcement learning for continuous control tasks,'' in \emph{Proceedings of the International Conference on Neuromorphic Systems (To appear in)}.\hskip 1em plus 0.5em minus 0.4em\relax IEEE, 2024.

\bibitem{b26}
M.~Towers, A.~Kwiatkowski, J.~Terry, J.~U. Balis, G.~De~Cola, T.~Deleu, M.~Goul{\~a}o, A.~Kallinteris, M.~Krimmel, A.~KG \emph{et~al.}, ``Gymnasium: A standard interface for reinforcement learning environments,'' \emph{arXiv preprint arXiv:2407.17032}, 2024.

\bibitem{b22}
E.~Todorov, T.~Erez, and Y.~Tassa, ``{MuJoCo:} a physics engine for model-based control,'' in \emph{2012 IEEE/RSJ International Conference on Intelligent Robots and Systems}.\hskip 1em plus 0.5em minus 0.4em\relax IEEE, 2012, pp. 5026--5033.

\end{thebibliography}

\end{document}